\pdfoutput=1

\documentclass[11pt]{article}

\usepackage{acl}

\usepackage{times}
\usepackage{latexsym}
\usepackage{graphicx}
\usepackage{enumitem}
\usepackage{subfig}
\usepackage{booktabs}
\usepackage{multirow}
\usepackage{hyperref}

\usepackage[T1]{fontenc}

\usepackage[utf8]{inputenc}

\usepackage{microtype}

\title{Harnessing Abstractive Summarization for Fact-Checked Claim Detection}

\author{Varad Bhatnagar\textsuperscript{1}, Diptesh Kanojia\textsuperscript{2,3}, Kameswari Chebrolu\textsuperscript{1}\\
 \textsuperscript{1}Department of Computer Science and Engineering, IIT Bombay, India \\
  \textsuperscript{2}Surrey Institute for People-Centred AI, \textsuperscript{3}Department of Computer Science \\
  \textsuperscript{2,3}University of Surrey, United Kingdom \\
  \texttt{\textsuperscript{1}\{varadhbhatnagar,chebrolu\}@cse.iitb.ac.in}\\ \texttt{\textsuperscript{2,3}d.kanojia@surrey.ac.uk} \\}

\begin{document}
\maketitle
\begin{abstract}
Social media platforms have become new battlegrounds for anti-social elements, with misinformation being the weapon of choice. Fact-checking organizations try to debunk as many claims as possible while staying true to their journalistic processes but cannot cope with its rapid dissemination. We believe that the solution lies in partial automation of the fact-checking life cycle, saving human time for tasks which require high cognition. We propose a new workflow for efficiently detecting previously fact-checked claims that uses abstractive summarization to generate crisp queries. These queries can then be executed on a general-purpose retrieval system associated with a collection of previously fact-checked claims. We curate an abstractive text summarization dataset comprising noisy claims from Twitter and their gold summaries. It is shown that retrieval performance improves 2x by using popular out-of-the-box summarization models and 3x by fine-tuning them on the accompanying dataset compared to verbatim querying. Our approach achieves Recall@\textit{5} and MRR of 35\% and  0.3, compared to baseline values of 10\% and 0.1, respectively. Our dataset, code, and models are
available publicly \href{https://github.com/varadhbhatnagar/FC-Claim-Det/}{here}.
\end{abstract}

\section{Introduction}
\label{sec:introduction}

Social media is increasingly used for business, entertainment, and political discourse, thus, encouraging users to produce and consume large volumes of information that may not always be accurate. Due to a lack of digital awareness, the masses often believe and forward such disputed claims in their social circles. Such spread of misinformation often culminates in incidents which cause damage to life and property. It is well documented that misinformation is used as a tool by political agents to slander their opposition~\citep{NBERw23089} and influence the opinion of the masses. It becomes furthermore dangerous when such claims pertain to religious beliefs, often leading to violence and mob lynchings\footnote{\href{https://www.washingtonpost.com/politics/2020/02/21/how-misinformation-whatsapp-led-deathly-mob-lynching-india/}{Article on Mob Lynching: Washington Post}}. In the era of COVID-19, unverified medical advice has also been circulated on social media \citep{SHAHI2021100104} which has already led to various health hazards.

Social media platforms have undertaken concerted efforts to tackle the fake news epidemic by enforcing strict policies to weed out unverified and sensitive content and ban habitual offenders. Journalistic organizations such as Alt News\footnote{\href{https://www.altnews.in/}{AltNews: Website}}, Factly\footnote{\href{https://factly.in/}{Factly: Website}}, Boom Live\footnote{\href{https://www.boomlive.in/}{Boom Live: Website}} and Snopes\footnote{\href{https://www.snopes.com/}{Snopes: Website}} among others are also fighting this problem by publishing fact-checking articles investigating the veracity of viral dubious claims. These articles detail the journalistic procedures followed to fact-check the claim along with suitable references.

Numerous researchers are working on AI-based solutions for fact-checking claims. Many datasets~\citep{thorne-etal-2018-fever, sathe-etal-2020-automated, fan-etal-2020-generating, schuster-etal-2021-get} have been released to train models which can automate sub-tasks such as claim verification, evidence retrieval and assigning a verdict in a fact-checking workflow. A critical and insufficiently researched step in the fact-checking workflow is- \textit{detecting whether a claim has been fact-checked previously}. This is a repetitive task with immense scope for automation, shrinking the turnaround time for a claim and ensuring that human efforts are put to better use on tasks involving higher cognition, such as assigning a verdict. In literature, learning-to-rank models \citep{shaar-etal-2020-known, vo-lee-2020-facts, mansour2022did} have been proposed for this step, which on being queried, produce a ranked list of results from a closed dataset of previously verified claims. Dozens of fact-checking articles are being published every hour around the world. It is difficult for journalistic organizations to maintain such a large real-time collection of fact-checked articles and claims, thus, making such an approach infeasible in real-world scenarios.

\begin{figure}
    \includegraphics[width=\linewidth]{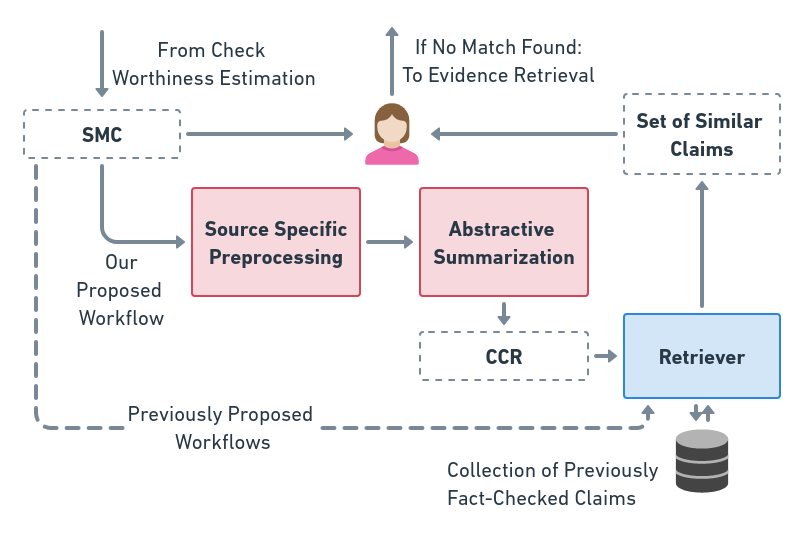}
    \caption{\textbf{Proposed Workflow}. In this work, we use Fact Check Explorer as a black box Retriever. The workflow proposed by previous works is denoted by a dotted path.}
    \label{fig: workflow}
    \vspace{-0.5cm}
\end{figure}

We propose a novel workflow (as shown in Figure~\ref{fig: workflow}) for detecting previously fact-checked claims. In this work, we use Google's \textit{Fact Check Explorer}, a cross-publisher, cross-language search engine for previously fact-checked articles, as a black box retriever. As social media platforms contribute a great deal to spreading misinformation, we deal with naturally occurring textual claims on Twitter in this study as opposed to artificial, well defined and structured claims~\citep{thorne-etal-2018-fever, aly-etal-2021-fact}. Instead of querying using verbatim claims, which are noisy, it is proposed that abstractive text summarization be used as a precursor to querying to generate clear, succinct queries capturing the claim in a minimum number of words. Figure~\ref{fig: workflow} represents a subpart of the complete fact-checking pipeline \citep{barron2020checkthat} with our proposed changes in red. \textit{CCR} and \textit{SMC} are defined in Section~\ref{sec:dataset}. In literature, no dataset exists for abstractive summarization of tweets, and no attempts have been made to address this problem using the \textit{Fact Check Explorer} to the best of our knowledge. Our contributions can be distilled into the following:
\begin{enumerate}[topsep=0pt,itemsep=-1ex,partopsep=1ex,parsep=1ex]
    \item \textbf{Workflow}: A novel workflow for detecting previously-fact checked claims at scale.
    \item \textbf{Dataset}: An abstractive summarization dataset\footnote{Data and models are made available here: \url{https://github.com/varadhbhatnagar/FC-Claim-Det/}} for tweets in the Indian context.
    \item \textbf{Models}: Popular and large pre-trained abstractive summarization models, fine-tuned under supervision on this data, which can be used for other purposes involving tweet summarization.
    \item \textbf{Experimental Study and Analysis}: We also perform quantitative and qualitative analysis for various outputs in our proposed workflow, including an analysis of generated summaries.
\end{enumerate}

The rest of this paper is organized as follows: Section~\ref{sec:relw} discusses related work, Section~\ref{sec:dataset} presents the dataset, Section~\ref{sec:approach} discusses the approach, evaluation metrics and the experimental setup. The results are presented and analysed in Section~\ref{sec:result} followed by Section~\ref{sec:conclu} which concludes the work and proposes future research directions. Section~\ref{sec:ethics} and \ref{sec:limitations} discuss the ethical considerations and limitations.

\section{Related Work}
\label{sec:relw}

\citet{sharma2019combating} describe the menace of misinformation on the Internet and summarize mitigation techniques and available datasets in this domain. Available intelligent technologies to assist the process of fact-checking are surveyed by~\citet{Nakov2021AutomatedFF}. This work highlights the partial overlap between current research endeavours and fact-checkers desiderata over the life cycle of a claim in a fact-checking pipeline. 

A general-purpose four-step automatic fact-checking pipeline is presented by~\citet{barron2020checkthat}. The task of determining if a claim has been previously fact-checked is the second step in the pipeline. This problem is addressed in a series of open challenges~\cite{clef-checkthat:2021:task2} at \textit{Checkthat!} workshop~\citep{barron2020checkthat, nakov2021clef, nakov2022clef} as part of CLEF \footnote{\href{http://www.clef-initiative.eu/}{CLEF: Website}}.
\citet{shaar-etal-2020-known} collect, annotate and release datasets of claim pairs and evidence sets, sourced from Politifact\footnote{\href{https://www.politifact.com/}{Politifact: Website}} and Snopes for solving this task. They develop and demonstrate the robustness of BM25 and BERT~\citep{devlin-etal-2019-bert} based learning to rank models on their dataset for this task. \citet{vo-lee-2020-facts,mansour2022did} also propose variants of a ranking approach to solve this problem. Further,~\citet{shaar2021role} work with data from political debates and model the context of a claim and illustrate the positive impact this has in determining if it has been previously fact-checked.
\citet{shaar2021assisting} publish a dataset and develop a system for detecting all previously fact-checked claims in a lengthy document. 

Text summarization has been used to enable verdict explainability in automatic fact-checking~\citep{mishra2020generating, stammbach2020fever} but it hasn't been used for denoising tweets, to the best of our knowledge.

\citet{tchechmedjiev2019claimskg} publish the \textit{ClaimsKG} Knowledge Graph, containing 28K fact-checked claims and their metadata such as sources, truth value and entities. Structured queries can be executed on this knowledge graph, enabling exploration and information discovery. However, it does not provide any mechanism to check if a claim has been previously fact-checked. \textit{Fact Check Explorer}\footnote{\href{https://toolbox.google.com/factcheck/explorer}{Fact Check Explorer: Web Search}} is a tool developed by Google which provides browsing and searching capability for already fact-checked articles which have the ClaimReview Schema\footnote{\href{https://schema.org/ClaimReview}{ClaimReview Schema}} embedded. There are performance limitations associated with this tool in the face of long and complex queries. 

\section{Dataset}
\label{sec:dataset}
The following terms are defined for lucid perusal of this work:

\begin{enumerate}[topsep=0pt,itemsep=-1ex,partopsep=1ex,parsep=1ex]
    \item \textbf{Social Media Claim (SMC):} A social media post (tweet, in this work) containing a claim in need of fact-checking. It is analogous to the output of the first step (\textit{check worthiness estimation}) in the automatic fact-checking pipeline presented by~\citet{barron2020checkthat}.
    \item \textbf{Fact Checked Article (FCA):} An article published by a fact-checking organization accepting or refuting a claim\footnote{\href{https://www.altnews.in/scene-from-2017-movie-viral-as-actual-shark-attack-on-helicopter/}{Example FCA}}.
    \item \textbf{Summary of Claim Review (SCR):} A short summary of the claim added by the publishing organization as part of the ClaimReview Schema associated with every FCA. Our use of the term SCR is the same as \textit{VerClaim} coined by~\citet{shaar-etal-2020-known}. 
    \item \textbf{Condensed Claim Representation (CCR):} A summary of the SMC generated using trained models.
\end{enumerate}

\subsection{Dataset Curation}
\label{sec:dataset_curation}
In this work, we focus on FCAs published by Indian organizations between 2018 and 2022. FCAs from the following IFCN\footnote{\href{https://www.poynter.org/ifcn/}{IFCN: Website}} certified organizations: 1) Alt News, 2) BoomLive, 3) India Today, 4) The Logical Indian, 5) The Quint, 6) Factchecker, 7) FactCrescendo, 8) Vishwas News, 9) PolitiFact, 10) Snopes, and 11) Factcheck.org, were retrieved. In order to make our dataset diverse, some FCAs from the USA based fact-checkers are also included, which shows that this workflow can be generalized. 

Twint\footnote{\href{https://github.com/twintproject/twint}{Twint: Github Repository}} is used to crawl Twitter, looking for URLs of the organizations mentioned above, in the comment threads of tweets. This resulted in a coarse collection of \textit{<Tweet, SCR>} pairs. Those pairs with tweets in languages other than English and tweets containing only image/video content are discarded. We perform annotation on this collection, keeping two aspects in mind: (1) the tweet should contain a claim, and (2) it should be textually summarizable to the corresponding SCR. URL removal from SMCs followed by pairwise de-duplication is performed at this stage, resulting in our final dataset, a collection of \textit{<SMC, SCR>} pairs, which can be used for training abstractive text summarization models. \textit{The final dataset only contains \textit{<SMC,SCR>} pairs where both are in English.}

Key world and Indian events have been covered as part of this dataset, such as the onset of COVID-19 and subsequent immunisation, the Taliban takeover of Afghanistan, Indian General Elections 2019 and US Presidential Elections 2020. Our annotation process is detailed below.

\subsubsection{Annotation Details}
\label{sec:annotation}

Two trained annotators were tasked with annotating every \textit{<Tweet,SCR>} pair from the coarse collection (Subsection~\ref{sec:dataset_curation}). Three categorical attributes viz. Tweet language, SCR language, category and one boolean attribute \textit{viz.} `Summarizability' had to be populated for each pair. 

The annotators were provided instructions to mark a pair as `summarizable' only when the SCR is a condensed version of the tweet and named entity coverage is more than 50\%. For deciding entity coverage, the annotators were allowed to take cues from the mentions and hashtags in the tweet. As majority of the FCA Publishers we dealt with are Indian, a lot of tweets and SCRs were in Indian languages such as Hindi, Hindi transliterated in English, Tamil, Telugu, and some other Indian languages. Any such instances were pruned from our dataset. 

To understand the motivation behind these SMCs, our annotators were also requested to categorize them into classes like a) Politics, b) Crime and Terrorism, c) World, d) Entertainment, e) Technology, f) Food, g) Religion, h) Sports, i) Health, j) Education, k) Business, l) Environment, and m) Other (miscellaneous). Though not relevant to this work, nor a part of the final dataset, we collect and annotate this data as well for further research.

We observe an inter-annotator agreement of 92\% within the annotations provided by both. 

\subsection{Dataset Statistics}
\label{sec:dataset_stats}

\begin{table}[h!]
\centering
\resizebox{0.45\textwidth}{!}{%
\begin{tabular}{@{}lcll@{}}
\toprule
\textbf{Data Entity}        & \multicolumn{2}{c}{\textbf{Count}}                  &               \\ \midrule
\textit{<SMC,SCR>} pairs             & \multicolumn{2}{c}{567}                             &               \\
Unique SMC                  & \multicolumn{2}{c}{531}                             &               \\
Unique SCR                  & \multicolumn{2}{c}{369}                             &               \\ \midrule
\textbf{FCA Source Country} & \multicolumn{2}{c}{\textbf{}}                       &               \\ \midrule
India                       & \multicolumn{2}{c}{93\%}                             &               \\
US                          & \multicolumn{2}{c}{7\%}                              &               \\ \midrule
\textbf{Median Length}      & \multicolumn{1}{l}{\textbf{Chars}} & \textbf{Words} &               \\ \midrule
SMC                         & \multicolumn{1}{l}{193}            & 33             &               \\
SCR                         & \multicolumn{1}{l}{70}             & 11             &               \\ \midrule
\textbf{Data Sets} & \multicolumn{3}{c}{\textbf{\begin{tabular}[c]{@{}c@{}}Cosine Similarity\\ Threshold\end{tabular}}} \\
\textbf{}                   & \multicolumn{1}{l}{\textbf{0.25}}  & \textbf{0.5}   & \textbf{0.75} \\ \midrule
NP                          & \multicolumn{1}{l}{59\%}            & 12\%            & 2\%            \\
P-H-M                       & \multicolumn{1}{l}{61\%}            & 13\%            & 3\%            \\
Snopes \citep{shaar-etal-2020-known}                      & \multicolumn{1}{l}{50\%}            & 8\%             & 1\%            \\ \bottomrule
\end{tabular}}
\caption{\textbf{Dataset Statistics and Complexity Analysis}. NP and P-H-M are defined in Subsection \ref{preprocessing}.}
\label{table:data_statistics}
\end{table}

\begin{figure}[ht!]%
    \centering
    \subfloat[\centering FCA Sources ]{{\includegraphics[width=3.4cm]{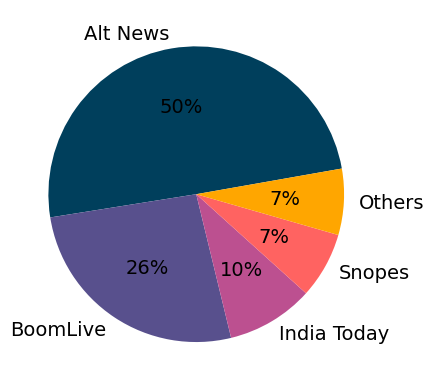} }}%
    \subfloat[\centering SMC Topics ]{{\includegraphics[width=4cm]{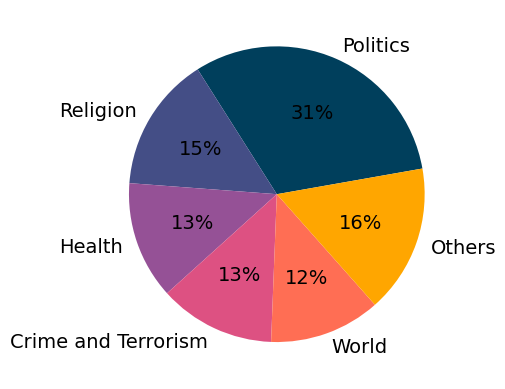} }}%
    \caption{\textbf{Dataset Distribution}}%
    \label{fig:dataset_distribution}%
    \label{f1}
\end{figure}

The statistics of our final dataset, comprising of 567 unique \textit{<SMC, SCR>} pairs are presented in Table~\ref{table:data_statistics}. Owning to several tweets and several FCAs about the same underlying event, 1:1 correspondence is not observed in the dataset, as evident from the first section in this table. Due to the 280 character limit imposed on tweets by Twitter, the SMCs are not arbitrarily long, with a median length of 33 words and the SCRs are observed to be very short, with a median length of 11 words. Similar to~\cite{shaar-etal-2020-known}, the complexity of the task is analyzed by reporting the word-level TF-IDF weighted cosine similarity for \textit{<SMC, SCR>} pairs. Since our dataset supports summarization, cosine similarity is higher compared to the Snopes dataset by~\cite{shaar-etal-2020-known}, as expected. Figure~\ref{fig:dataset_distribution} presents the FCA Source and SMC Topic distribution. 46\% of the SMCs are political or religious, which is no surprise as these sensitive topics polarise opinion very easily. A large chunk of SMCs are health-related, owning to misinformation surrounding the COVID-19 immunization and mass hysteria.

\section{Our Approach}
\label{sec:approach}

SMCs are very noisy in nature due to the inherent way people interact on social media and micro-blogging platforms. On Twitter, tweets are bounded by a character limit, forcing people to use slang and abbreviations to communicate effectively. It also allows for mentions and hashtags to be embedded in tweets to encourage inter-user interaction. Using these noisy SMCs verbatim (as done by~\citet{shaar-etal-2020-known}) to check if they have been previously fact-checked is challenging, as the retrieval module has to do all the heavy lifting for which it is not equipped. 

In this work, it is hypothesized that a system which extracts queryable content from SMCs by dealing with its syntactic and semantic aspects before querying the retrieval module should perform better than verbatim querying. Keeping in mind the small scale at which fact-checking organizations work and the continuously growing collection of FCAs, Google's \textit{Fact Check Explorer} is used as a retriever for previously fact-checked claims instead of a closed collection of verified claims. The \textit{Fact Check Explorer} indexes the latest FCAs across the world and provides easy to use search APIs for free, which support filtering based on publisher and language, among other features. Various text pre-processing techniques on SMCs are experimented with before using state-of-the-art abstractive text summarization models to generate corresponding CCRs. These CCRs are then used to query the retriever. These techniques and the models used are detailed in the following subsections. Our preference for abstractive summarization over extractive summarization arises because of two reasons; the SMCs are noisy and unlikely to contain query-able spans and due to the recent progress in abstractive summarization research~\citep{lewis2019bart,zhang2020pegasus,JMLR:v21:20-074,10.5555/3491440.3491993}.

This proposed workflow is generic in nature as \textit{<SMC, SCR>} pairs collected from other microblogging platforms (using our curation methodology) can be used to train summarization models after applying text pre-processing techniques specific to that platform. These models can also generate queries for open-domain evidence retrieval, which is the next step in a fact-checking pipeline. It is also futuristic in the sense that a generative module can replace the text summarization module to support multimodal SMCs. However, this work is kept limited to textual SMCs due to a lack of suitable labelled data and the absence of a reliable equivalent of \textit{Fact Check Explorer} for joint text, image and video search.

\subsection{Twitter Specific Preprocessing}
\label{preprocessing}

Most social media platforms encourage inter-user interaction by allowing 'mentioning' other users in a post. Typically, some form of notification goes to the user being mentioned, getting his attention on the post content. It is also used as a way for tagging people to establish their presence in photos and videos. Hashtags are metadata tags which allow cross referencing of content by topic or theme. They typically identify with some event or social movement, allowing users to discover and associate with trending content. Both hashtags and mentions are available on Twitter along with emojis, which are smileys embedded in text, providing emotional cues to the reader. 

We experiment with these three aspects of a tweet. Most search engines do not deal with Emojis, hence we replace them with a constant to form the P+MRep set. Hashtags and Mentions provide rich signals about named entities, hence it is important to incorporate them in the input in some way. Upon manual analysis of the data, it was seen that a lot of tweets mentioned users who were unrelated to the content in the tweet. Some recurring instances of this phenomena that we came across, were fact-checking requests mentioning many journalists and organizations and political tweets mentioning prominent members of the opposition political party and prominent believers of the opposite ideology. It was observed that hashtags were also used in a similar manner. Another observation was the existence of runs of space separated hashtags and mentions and their occurrence at the beginning or end of the tweet, signifying the preference of users to separate actual tweet content from these meta tags. These signals led us to create sets of data where mention and hashtag runs are removed except the first member in each run (P-MRR-HRR). Further, some users used organization related twitter handles or twitter handles in other languages like Hindi. To deal with this, we replace these by their original names on Twitter to get the P-MRR-HRR+MRep set. With a clear intuition behind such preprocessing, we now describe what Twitter specific text preprocessing techniques are applied to SMCs to produce the following \textit{<SMC,SCR>} sets, from the final dataset:
\begin{enumerate}[topsep=0pt,itemsep=0ex,partopsep=1ex,parsep=1ex]
    \item \textbf{Verbatim (NP)}: SMCs are used verbatim.
    \item \textbf{Preprocessed (P)}: Symbols for hashtags(\textit{\#}) and mentions(\textit{@}), emojis, punctuation and redundant are removed, followed by lowercasing of SMCs.
    \item \textbf{Pre-processed with Emojis Replaced  (P+ERep)}: Emojis are replaced by the string \textit{\$EMOJI\$} in addition to techniques used in P.
    \item \textbf{Pre-processed with Hashtags and Mentions Removed (P-H-M)}: All hashtags and mentions are removed in addition to techniques used in P. Subsets with only hashtag removal \textbf{(P-H)} and only mention removal \textbf{(P-M)} are also created.
    \item \textbf{Pre-processed with Mention and Hashtag Run Removed (P-MRR-HRR)}: Run of hashtags and mentions are removed, except the first entity in each run, in addition to techniques used in P.
    \item \textbf{Pre-processed with Mention and Hashtag Run Removed and Mentions Replaced (P-MRR-HRR+MRep)}: The remaining mentioned handles in P-MRR-HRR are replaced by their official names from Twitter.
\end{enumerate}

\subsection{Summarization Models}
For summarization, the following models were experimented with:
\label{models}
\begin{enumerate}[topsep=0pt,itemsep=-1ex,partopsep=1ex,parsep=1ex]
    \item \textbf{Truncate {\textit{k}}}: A naive summarizer which truncates a SMC to the first \textit{k} space-separated tokens. It is used as a baseline to show gains by more complex models.
    \item \textbf{T5}: A transformer-based architecture by~\citet{JMLR:v21:20-074} that uses a text to text approach for all tasks. It is pre-trained on a multi-task mixture of supervised and unsupervised tasks such as denoising on the high quality C4 corpus, sentiment analysis, natural language inference and question answering, among others. To make the model cope with this multi-task training, a task-specific prefix is added to the input sentence.
    \item \textbf{BART}: A transformer-based sequence to sequence model by~\citet{lewis2019bart} which incorporates the bidirectional encoder of BERT~\cite{devlin-etal-2019-bert} and the left-to-right autoregressive decoder of GPT~\cite{Radford2018ImprovingLU, radford2019rewon}, pre-trained in denoising autoencoder style. It works well for downstream tasks involving text generation.
    \item \textbf{PEGASUS}: A transformer-based sequence to sequence model by~\citet{zhang2020pegasus} which uses a self-supervised pre-training objective called gap-sentence generation, aimed at optimizing downstream abstractive summarization tasks. In gap-sentence generation, important sentences in a document are masked, and the transformer model is asked to predict those sentences. PEGASUS shows impressive performance even with a small number of samples during fine-tuning.
\end{enumerate}

\subsection{Decoding Strategies}
Decoding strategies define how text should be generated by models that support language generation. Based on the end application, the model may be expected to generate text that can be lengthy, short, non-repetitive, interesting, surprising, and so on. Our application requires the output to be short and crisp. In this work, we experiment with Greedy, Beam Search, Top \textit{k} and Top \textit{p}~\citep{holtzman2019curious} decoding strategies.

\subsection{Evaluation Metrics}
\begin{figure}[h!]%
    \centering
    \includegraphics[width=0.9\linewidth]{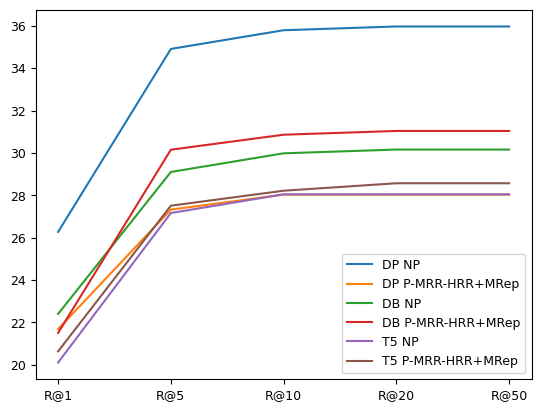}
    \label{f2}
    \caption{\textbf{Recall Plateauing for Decoding Strategies}}
    \label{fig: recall_plat}
\end{figure}

Recall@\textit{k} and Mean Reciprocal Rank (MRR) are reported for all experiments, as is the norm in retrieval tasks. While checking if a claim was previously fact-checked, fact-checkers would not want to look beyond the first few results. Keeping this in mind, Recall@\textit{5} is used as the primary metric for comparing retrieval performance. Figure~\ref{fig: recall_plat} shows the variation in Recall@\textit{k} with increasing value of \textit{k}. The sharp bend at Recall@\textit{5}, subsequent plateauing also motivated us to report this metric for retrieval. Also, it is practically feasible for a human fact-checker to go through 5 results per claim rather than 10 or 20 results.

For evaluating the quality of the summary generated, word-level TF-IDF weighted cosine similarity between SMCs and CCRs and between SCRs and CCRs is reported. BLEU4~\citep{papineni-etal-2002-bleu}.

\begin{table*}[t]
\resizebox{\textwidth}{!}{%
\begin{tabular}{@{}lrrrrrrrrrrrrrrrrl@{}}
\multicolumn{1}{l|}{} &
  \multicolumn{4}{c|}{\textbf{No Summarization}} &
  \multicolumn{6}{l|}{\textbf{Summarization using Out of the Box Models}} &
  \multicolumn{6}{c}{\textbf{Summarization using Fine Tuned Models}} &
   \\ \cmidrule(r){1-17}
\multicolumn{1}{l|}{\multirow{2}{*}{\textbf{\begin{tabular}[c]{@{}l@{}}Preprocessing\\ Strategies\end{tabular}}}} &
  \multicolumn{2}{c}{\textbf{None}} &
  \multicolumn{2}{c|}{\textbf{Truncate11}} &
  \multicolumn{2}{c}{\textbf{T5}} &
  \multicolumn{2}{c}{\textbf{D BART}} &
  \multicolumn{2}{c|}{\textbf{D PEGASUS}} &
  \multicolumn{2}{c}{\textbf{T5}} &
  \multicolumn{2}{c}{\textbf{D BART}} &
  \multicolumn{2}{c}{\textbf{D PEGASUS}} &
   \\
\multicolumn{1}{l|}{} &
  \multicolumn{1}{c}{\textbf{R@5}} &
  \multicolumn{1}{c}{\textbf{MRR}} &
  \multicolumn{1}{c}{\textbf{R@5}} &
  \multicolumn{1}{c|}{\textbf{MRR}} &
  \multicolumn{1}{c}{\textbf{R@5}} &
  \multicolumn{1}{c}{\textbf{MRR}} &
  \multicolumn{1}{c}{\textbf{R@5}} &
  \multicolumn{1}{c}{\textbf{MRR}} &
  \multicolumn{1}{c}{\textbf{R@5}} &
  \multicolumn{1}{c|}{\textbf{MRR}} &
  \multicolumn{1}{c}{\textbf{R@5}} &
  \multicolumn{1}{c}{\textbf{MRR}} &
  \multicolumn{1}{c}{\textbf{R@5}} &
  \multicolumn{1}{c}{\textbf{MRR}} &
  \multicolumn{1}{c}{\textbf{R@5}} &
  \multicolumn{1}{c}{\textbf{MRR}} &
  \multicolumn{1}{c}{} \\ \cmidrule(r){1-17}
\multicolumn{1}{l|}{\textbf{NP}} &
  9.52 &
  .09 &
  17.28 &
  \multicolumn{1}{r|}{.14} &
  15.52 &
  .13 &
  20.46 &
  .17 &
  \textbf{22.40} &
  \multicolumn{1}{r|}{\textbf{.19}} &
  27.16 ±2.55 &
  .23 ±.02 &
  29.10 ±3.15 &
  .26 ±.02 &
  \textbf{34.91 ±5.91} &
  \textbf{.30 ±.05} &
   \\ \cmidrule(r){1-17}
\multicolumn{1}{l|}{\textbf{P}} &
  11.99 &
  .11 &
  18.34 &
  \multicolumn{1}{r|}{.15} &
  17.64 &
  .15 &
  17.99 &
  .14 &
  21.52 &
  \multicolumn{1}{r|}{.17} &
  28.38 ±6.55 &
  .24 ±.05 &
  28.21 ±6.88 &
  .24 ±.06 &
  27.69 ±2.63 &
  .24 ±.02 &
   \\
\multicolumn{1}{l|}{\textbf{-H}} &
  12.70 &
  .12 &
  17.99 &
  \multicolumn{1}{r|}{.15} &
  17.28 &
  .15 &
  17.64 &
  .14 &
  20.46 &
  \multicolumn{1}{r|}{.16} &
  24.87 ±5.13 &
  .21 ±.04 &
  30.15 ±6.08 &
  .26 ±.06 &
  29.61 ±6.59 &
  .25 ±.05 &
   \\
\multicolumn{1}{l|}{\textbf{-M}} &
  13.05 &
  .12 &
  18.69 &
  \multicolumn{1}{r|}{.15} &
  18.34 &
  .15 &
  17.64 &
  .14 &
  21.16 &
  \multicolumn{1}{r|}{.17} &
  26.80 ±5.33 &
  .23 ±.03 &
  \textit{30.15 ±4.98} &
  .26 ±.04 &
  29.10 ±2.45 &
  .25 ±.02 &
   \\
\multicolumn{1}{l|}{\textbf{-H-M}} &
  13.93 &
  .13 &
  \textbf{18.87} &
  \multicolumn{1}{r|}{\textbf{.15}} &
  17.81 &
  .15 &
  17.28 &
  .14 &
  20.28 &
  \multicolumn{1}{r|}{.16} &
  26.46 ±4.69 &
  .23 ±.03 &
  27.51 ±4.09 &
  .23 ±.04 &
  26.28 ±2.36 &
  .23 ±.02 &
   \\
\multicolumn{1}{l|}{\textbf{+ERep}} &
  10.58 &
  .10 &
  17.46 &
  \multicolumn{1}{r|}{.14} &
  16.58 &
  .14 &
  17.99 &
  .15 &
  22.05 &
  \multicolumn{1}{r|}{.18} &
  27.51 ±6.13 &
  .23 ±.05 &
  28.20 ±5.89 &
  .25 ±.05 &
  30.15 ±5.72 &
  \textit{.27 ±.05} &
   \\
\multicolumn{1}{l|}{\textbf{-MRR-HRR}} &
  12.35 &
  .11 &
  17.81 &
  \multicolumn{1}{r|}{.15} &
  17.46 &
  .15 &
  17.81 &
  .14 &
  21.69 &
  \multicolumn{1}{r|}{.18} &
  28.75 ±6.38 &
  .25 ±.05 &
  27.33 ±5.94 &
  .23 ±.05 &
  25.92 ±1.98 &
  .22 ±.01 &
   \\
\multicolumn{1}{l|}{\textbf{\begin{tabular}[c]{@{}l@{}}-MRR-HRR \\ +MRep\end{tabular}}} &
  12.70 &
  .12 &
  \textbf{18.87} &
  \multicolumn{1}{r|}{\textbf{.15}} &
  18.34 &
  .15 &
  17.81 &
  .14 &
  21.87 &
  \multicolumn{1}{r|}{.18} &
  27.51 ±5.43 &
  .24 ±.04 &
  30.15 ±5.98 &
  .25 ±.05 &
  27.32 ±3.87 &
  .24 ±.03 &
   \\ \cmidrule(r){1-17}
\textbf{Skyline} &
  \textbf{63.85} &
  \textbf{.55} &
  \multicolumn{1}{l}{} &
  \multicolumn{1}{l}{} &
  \multicolumn{1}{l}{} &
  \multicolumn{1}{l}{} &
  \multicolumn{1}{l}{} &
  \multicolumn{1}{l}{} &
  \multicolumn{1}{l}{} &
  \multicolumn{1}{l}{} &
  \multicolumn{1}{l}{} &
  \multicolumn{1}{l}{} &
  \multicolumn{1}{l}{} &
  \multicolumn{1}{l}{} &
  \multicolumn{1}{l}{} &
  \multicolumn{1}{l}{} &
   \\ \cmidrule(r){1-17}
\end{tabular}}
\caption{\textbf{Retrieval Results} (Subsection \ref{sec:result_retrieval}). D BART and D PEGASUS stand for Distilled BART and Distilled PEGASUS respectively, and  Recall@\textit{5} is represented by R@5. All preprocessing strategies prefixed with '+' or '-' are applied on top of the P set.}
\label{table: retrieval_results}
\end{table*}
\subsection{Experiment Setup}
\label{sec:exp_setup}
In the experiments, the performance of summarization models in both out-of-the-box settings and through fine-tuning, \textit{i.e.,}  training on the task under supervision are compared. For the fine-tuning experiments, 5-fold cross-validation is performed on the data; and mean values along with the standard deviation are observed. Other experiments are performed on the entire data without any splits as no parameter learning is involved.

All experiments performed with the help of Transformer-based architectures in Table~\ref{table: retrieval_results} use Beam Search decoder with a beam size of 6 and the maximum token length of a generated sequence, set to 15 with early-stopping enabled. For Truncate \textit{k} experiments, \textit{k}=11 is set. We arrive at these constants by looking at the median summary length provided in Table~\ref{table:data_statistics} and giving some leeway to transformer models as they operate on sub-word vocabularies. Hugging Face\footnote{\href{https://huggingface.co/}{Hugging Face: Website}} implementations of the models mentioned in Subsection~\ref{models} are used for all experiments involving transformers. For the PEGASUS and BART experiments, we use the distilled versions released by~\citet{shleifer2020pre} using the shrink and fine-tune approach on the CNN dataset. The 16-layer-encoder 4-layer-decoder version of distilled PEGASUS\footnote{\href{https://huggingface.co/sshleifer/distill-pegasus-cnn-16-4}{Distilled PEGASUS Model}} and 12-layer-encoder 6-layer-decoder version of distilled BART\footnote{\href{https://huggingface.co/sshleifer/distilbart-cnn-12-6}{Distilled BART Model}} are used. The base version of T5\footnote{\href{https://huggingface.co/t5-base}{T5-base Model}} is used for all experiments involving T5. The number of trainable parameters are 220M, 300M and 370M in T5-base, Distilled BART and Distilled PEGASUS, respectively. Since \textit{Fact Check Explorer} is an ever-growing and evolving system, CCRs are generated for all experiments first, and then retrieval queries are run, ensuring consistency across results. For retrieval, the API documentation\footnote{\href{https://developers.google.com/fact-check/tools/api/reference/rest/v1alpha1/pages}{Fact Check Explorer: API Documentation}} is followed and retrieved URLs are compared with normalized (removing redirection/parameters accompanying the URL) URLs associated with an FCA.

\section{Results and Discussion}
\label{sec:result}

This section discusses the results obtained at various stages of our workflow.

\subsection{Retrieval}
\label{sec:result_retrieval}

We present the retrieval results in Table~\ref{table: retrieval_results}. From top to bottom, the SMC pre-processing (Section~\ref{sec:dataset}) complexity increases and from left to right, the complexity of the summarization models (Subsection~\ref{models}) increases.
For the skyline numbers, \textit{Fact Check Explorer} is queried using the gold SCRs, giving 63.85 Recall@\textit{5} and 0.55 MRR. We observe two evident trends via experimentation- (1) the performance gain in using an out-of-the-box summarization model, as compared to no summarization, and (2) the benefit of learning under supervision on our labelled dataset, indicated by the sharp gain in performance of fine-tuned models as compared to the corresponding out of the box models. Since the PEGASUS model is pre-trained with an objective to boost abstractive summarization performance, it works quite well out-of-the-box, giving a 2x increase in performance compared to no summarization. The best performing model is Distilled PEGASUS, fine-tuned on our dataset (without any pre-processing), as exhibited by a Recall@\textit{5} of 34.91 and MRR of 0.3, which is more than 3x improvement over verbatim querying. 

We use three different summarization strategies- (1) No Summarization, (2) Summarization using out-of-the-box Models, and (3) Summarization using fine-tuned models as shown in Table~\ref{table: retrieval_results}. We separately highlight the best performance in the table itself in each of these cases. 
In the no summarization experiments, we observe that complex pre-processing techniques lead to a performance gain, as indicated by the best Recall@\textit{5} and MRR of 18.87 and 0.15 on dealing with mentions and hashtags (for both P-H-M and P-MRR-HRR+MRep). Among the out-of-the-box experiments, it is seen that Distilled PEGASUS comfortably outperforms T5 and Distilled BART, with the best Recall@\textit{5} and MRR being 22.4 and 0.19, respectively. Highly parameterized models like BART and PEGASUS do not benefit from input pre-processing. 

The gap between the skyline numbers and the best performing model can be attributed to the fact that most models are pre-trained on document level summarization datasets such as CNN/Daily Mail~\citep{nallapati-etal-2016-abstractive} and Huge News~\cite{zhang2020pegasus}. Hence, they struggle with summarizing short input text.

\subsection{Summarization Quality}
\label{sec:result_summ_quality}
\begin{table}[t!]
\centering
\resizebox{0.45\textwidth}{!}{%
\begin{tabular}{cccccc}
\hline
\multirow{3}{*}{\textbf{Experiment}} & \multirow{3}{*}{\textbf{n-Gram}} & \multicolumn{3}{c}{\textbf{Cosine Similarity}} & \multirow{2}{*}{\textbf{BLEU4}} \\
                               &   & \multicolumn{3}{c}{\textbf{Threshold}}       &       \\
                               &   & \textbf{0.25} & \textbf{0.5} & \textbf{0.75} &       \\ \hline
\multirow{2}{*}{E1 SMC vs CCR} & 1 & 80\%          & 26\%         & 2\%           & -     \\
                               & 2 & 83\%          & 30\%         & 2\%           & -     \\
\multirow{2}{*}{E1 SCR vs CCR} & 1 & 76\%          & 38\%         & 14\%          & 39.7 \\
                               & 2 & 73\%          & 38\%         & 13\%          & 39.3 \\ \hline
\multirow{2}{*}{E2 SMC vs CCR} & 1 & 83\%          & 26\%         & 4\%           & -     \\
                               & 2 & 85\%          & 29\%         & 3\%           & -     \\
\multirow{2}{*}{E2 SCR vs CCR} & 1 & 68\%          & 31\%         & 11\%          & 38.9 \\
                               & 2 & 69\%          & 33\%         & 13\%          & 39.2 \\ \hline
\end{tabular}
}
\caption{\textbf{Summarization Quality Analysis} (Subsection \ref{sec:result_summ_quality}). E1 and E2 stand for Distilled PEGASUS with NP and P-MRR-HRR+MRep experiments respectively (as described in Section \ref{sec:approach}) and n-Gram stands for the value of \textit{n} in \textit{n}-grams not appearing more than once in beam search decoding.}
\label{table: summarization_quality}
\end{table}

The quality of CCRs is reported in Table~\ref{table: summarization_quality}. We compare CCRs with SMCs and SCRs on two metrics ~ (1) Word level TF-IDF Weighted Cosine Similarity and (2) BLEU4. Since BLEU4 is generally reported between reference and generated sequences, it does not make sense to report it for SMC vs CCR rows. \textit{We observe high BLEU4 scores for the CCRs} signifying that our approach can generate valid summaries, as can also be seen in Table~\ref{table: smc_SCR}. For both E1 and E2, our BLEU4 scores are approaching (approx.) 40. On comparing SMC vs CCR cosine similarities for both experiments with the cosine similarity of SMC vs SCR (last section in Table~\ref{table:data_statistics}), we find higher values for all thresholds indicating that \textit{our generated summaries are significantly similar to the tweets as compared to the gold summaries provided}.

\subsection{Decoding}
\label{sec:result_decoding}
\begin{figure}[ht!]%
    \centering
    \includegraphics[width=0.9\linewidth]{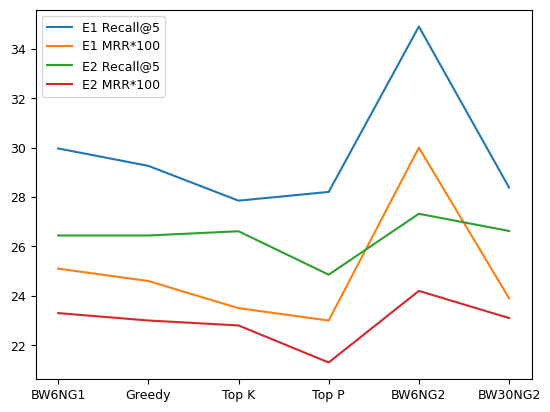}
    \caption{\textbf{Decoding Strategy Comparison} (Subsection \ref{sec:result_decoding}). MRR values are multiplied by 100 for better visualization. BW\textit{k}NG\textit{n} corresponds to beam search decoding with beam width \textit{k} and no \textit{n}-grams appearing more than once in the generated output. \textit{k} is set to 50 in Top \textit{k} and \textit{p} is set to 0.92 in Top \textit{p}.}
    \label{fig: decoding_strategy}
\end{figure}

\begin{table*}[t]
  \centering
  \resizebox{\textwidth}{!}{%
\begin{tabular}{llll}
\hline
\# &
  \textbf{SMC} &
  \textbf{SCR} &
  \textbf{CCR} \\ \hline
1 &
  \begin{tabular}[c]{@{}l@{}}Congratulations to Uttarakhand CM for \\ becoming the first CM ever to charge stranded\\  citizens for rescue operations! Helicopter \\ rides will now be chargeable during rescue \\ operations in Uttarakhand. And if you can't \\ pay, you may safely die. \#AchheDin \#BJP\end{tabular} &
  \begin{tabular}[c]{@{}l@{}}Passengers in Uttarkhand to be charged \\ for rescue operations\end{tabular} &
  \begin{tabular}[c]{@{}l@{}}Uttarakhand CM has charged stranded \\ citizens for helicopter rides during rescue\\  operations\end{tabular} \\ \hline
2 &
  \begin{tabular}[c]{@{}l@{}}@AltNews   We are getting various WhatsApp\\  forward regarding as Corona has been \\ emerges only due to 5G testing  in world.   \\ Please put some light,  seems ,it is only a \\ brain shit.\end{tabular} &
  \begin{tabular}[c]{@{}l@{}}5G radiation is the cause behind the \\ second wave of coronavirus pandemic \\ in India\end{tabular} &
  Coronavirus outbreak due to 5G testing \\ \hline
3 &
  \begin{tabular}[c]{@{}l@{}}This woman in Afghanistan was killed by Taliban\\  for not wearing the proper cloth. \#Afghanistan  \\ \#Taliban @cnn @FoxNews @BBCWorld\end{tabular} &
  \begin{tabular}[c]{@{}l@{}}Video shows a woman being shot in \\ the head by Taliban in Afghanistan \\ for not dressing appropriately\end{tabular} &
  \begin{tabular}[c]{@{}l@{}}Woman killed by Taliban for not wearing\\  proper cloth\end{tabular} \\ \hline
4 &
  \begin{tabular}[c]{@{}l@{}}"India is ranked 102nd in the global hunger index,\\  out of 117 countries. We are ranked in between\\  Niger \& Sierra Leone. We are the lowest ranked\\  South Asian country. Bangladesh is ranked 88th\\  and Pakistan 94th. They have only recently\\  overtaken us. Our rank was 55,only 5 years ago"\end{tabular} &
  \begin{tabular}[c]{@{}l@{}}India's ranking in Global Hunger Index\\  (GHI) has fallen from 55 in 2014 to 102\\  in 2019\end{tabular} &
  India ranked 102nd in the global hunger index \\ \hline
5 &
  \begin{tabular}[c]{@{}l@{}}"Oxygen donated  from Saudi and relabelled in\\  india by  Reliance, Share this with your contacts \\ in Saudi and make this viral .. Let the world know\\  the cheapness of this PM   "\end{tabular} &
  \begin{tabular}[c]{@{}l@{}}Oxygen sent from Saudi Arabia is being\\  distributed in the name of Reliance\end{tabular} &
  \begin{tabular}[c]{@{}l@{}}Reliance taking credit for oxygen supplied\\  by Saudi Arabia\end{tabular} \\ \hline
\end{tabular}}
  \caption{\textbf{SMCs and SCRs from the Dataset with corresponding CCRs} (Subsection \ref{sec:result_ccr_quality}).}
  \label{table: smc_SCR}
\end{table*}

Figure~\ref{fig: decoding_strategy} shows the variation in retrieval results on using different decoding strategies. Definitions for E1 and E2 follow from Table~\ref{table: summarization_quality}.

As observed from this figure, BW\textit{6}NG\textit{2} seems to be the best performing decoding strategy. Hence, this strategy is used for all experiments in Table~\ref{table: retrieval_results}. BW\textit{6}NG\textit{1} also seems to be a good alternative, but the \textit{1}-gram constraint makes the queries very terse and grammatically inconsistent (observed manually). Greedy, Top \textit{k} and Top \textit{p} strategies are not competitive for such a task.

\subsection{Larger Language Models}
\begin{figure}[ht!]%
\centering
    \includegraphics[width=0.9\linewidth]{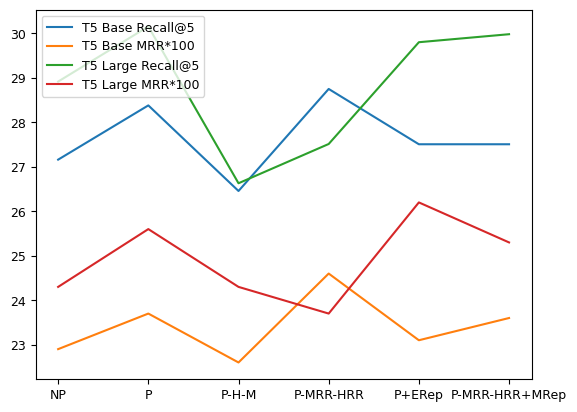}
    \caption{\textbf{Effect of Larger Language Models on Retrieval Metrics}}
    \label{fig: Larger LM}
\end{figure}

We study the variation in performance using even larger models such as T5 Large\footnote{\href{https://huggingface.co/t5-large}{T5-large Model}}, which has 770M parameters, three times that of T5-base. CCRs generated by the larger model perform better on both retrieval metrics across a variety of pre-processed SMCs, but the performance gain is not significant. It is offset by a longer training time and heavy compute requirements leading to considerable cost overheads. Since this is not a study of large generative models and given the modest resources owned by most fact-checking organizations, we do not explore any larger language models such as T5-3B and T5-11B, which have 3 Billion and 11 Billion parameters, respectively.

\subsection{CCR Quality}
\label{sec:result_ccr_quality}

Table~\ref{table: smc_SCR} lists a few CCRs generated by the best performing model, also listing the corresponding SMCs and SCRs. The model successfully extracts the core claim from the SMCs and ignores tokens like mentions and hashtags that have no contribution to the core claim. Owning to the constraints placed on length, it is seen that the generated CCRs are succinct and context-independent. They seem to be paraphrases of the gold SCRs, making them good candidates for querying the retrieval system. It is also seen that our model finds factual inputs which require reasoning, difficult to deal with. For instance, \#4 in Table~\ref{table: smc_SCR} requires a model to understand that going from rank 55 to 102 in the Global Hunger Index is a fall and not a rise. Our workflow does not expect the underlying language model to understand and reason, and this workflow only requires the generation of a valid summary.

\section{Conclusion and Future Work}
\label{sec:conclu}

In this work, a new workflow for detecting previously fact-checked claims is proposed. This workflow uses text summarization as an intermediate step before retrieval module invocation. Clean and crisp summaries thus generated are then used for querying a retrieval system. To this end, a first-of-its-kind tweet summarization dataset in the Indian context to train such models is curated and released under the \href{https://creativecommons.org/licenses/by-nc-sa/4.0/}{CC-BY-NC-SA 4.0 license}. The performance gained on using popular out-of-the-box and fine-tuned summarization models before querying the \textit{Fact Check Explorer} is demonstrated, and discussed with qualitative samples. Various popular decoding strategies are compared, and the implication of using larger pre-trained models is explored. \textit{The aim of this work is to aid in the creation of general-purpose and performant modules which can speed up a fact-checking pipeline by equipping fact-checkers with the tools to fight misinformation at a large scale.}

In future, we would also like to perform this task in a more general context for news items from various countries, extending our work in a multilingual scenario. Also, named entities are crucial in drafting a good query for any retrieval system. Generating summaries based on the Named Entities~\citep{zhang2020pegasus} found in SMCs is a promising avenue to explore. We do not take tweet threads into account as our focus is SMCs by users and not replies or comments to those SMCs, however, this can be an interesting future direction.  Other controlled text generation~\citep{keskar2019ctrl, chan2021cocon} techniques can also be explored to extract the maximum information from noisy SMCs.
Better pre-training objectives for abstractive summarization on noisy text can lead to efficient out-of-the-box models for this task.

Most Indian fact-checking organizations in Section~\ref{sec:annotation} also publish FCAs in regional languages such as Hindi, Tamil and Telugu. Twitter conversations, spreading misinformation in other pure and transliterated Indic languages are voluminous. Cross-lingual summarization research~\citep{zhu-etal-2019-ncls} would go a long way in fighting misinformation in a holistic manner.

\section{Acknowledgements}
We acknowledge the kind support provided by IMPRINT-2, a technology development initiative of MHRD and DST, Government of India.

\section{Ethical Considerations}
\label{sec:ethics}

To the best of our knowledge, no code of ethics was violated throughout the experiments performed for this study. We report all hyper-parameters and other technical details necessary to reproduce our results, and release the code and dataset curated via this work. We perform our experiments with the help of various language models which may contain biases as discussed by~\citet{DBLP:journals/corr/abs-2112-04359}. However, we believe that our workflow and methodology are solid and apply to any social media fake news setting. Any quantitative results reported by us are reproducible, subject to the ever growing number of articles indexed by the Fact Check Explorer (reported in Section~\ref{sec:exp_setup}). However, the qualitative results (like generated summaries) are an outcome of computational models that does not represent our personal views. We do not include any identifying information in the data that we use for our experiments and ensure that the dataset release will follow anonymization of any such information. 

We would like to state that this dataset is collected in a recent real-world setting (raw social media claims from 2018-2022) and \textit{no attempt has been made by us to subdue tweets on certain topics and promote others}. More precisely, we freely assigned the tweets to our annotators without any domain/topic specificity, however, they were required to label the tweet from a list of categories (Section~\ref{sec:annotation}) to collect more information.

\section{Limitations}
\label{sec:limitations}
We believe there is a limitation to our work,\textit{ i.e.,} \textbf{The limited size of this dataset}; which can be attributed to following reasons:
\begin{itemize}
    \item Most fact-checking organisations (covered in this work) emerged post-2017.
    \item Our data curation relies on a large number of users replying to potentially misinformative tweets. This user behaviour is limited by social network usage, awareness and internet proliferation for a particular language, region or country.
    \item  The ``manual pruning'' step while curating the tweet level summarization dataset was a very time/effort-intensive process. For e.g., around 5000 coarse \textit{<Tweet,SCR>} pairs were manually pruned to get the final dataset containing 567 \textit{<SMC,SCR>} pairs, implying a rejection rate close to 90\%.
\end{itemize}

\bibliography{anthology,custom}

\begin{thebibliography}{36}
\expandafter\ifx\csname natexlab\endcsname\relax\def\natexlab#1{#1}\fi

\bibitem[{Allcott and Gentzkow(2017)}]{NBERw23089}
Hunt Allcott and Matthew Gentzkow. 2017.
\newblock \href {https://doi.org/10.3386/w23089} {Social media and fake news in
  the 2016 election}.
\newblock Working Paper 23089, National Bureau of Economic Research.

\bibitem[{Aly et~al.(2021)Aly, Guo, Schlichtkrull, Thorne, Vlachos,
  Christodoulopoulos, Cocarascu, and Mittal}]{aly-etal-2021-fact}
Rami Aly, Zhijiang Guo, Michael~Sejr Schlichtkrull, James Thorne, Andreas
  Vlachos, Christos Christodoulopoulos, Oana Cocarascu, and Arpit Mittal. 2021.
\newblock \href {https://doi.org/10.18653/v1/2021.fever-1.1} {The fact
  extraction and {VER}ification over unstructured and structured information
  ({FEVEROUS}) shared task}.
\newblock In \emph{Proceedings of the Fourth Workshop on Fact Extraction and
  VERification (FEVER)}, pages 1--13, Dominican Republic. Association for
  Computational Linguistics.

\bibitem[{Barr{\'o}n-Cedeno et~al.(2020)Barr{\'o}n-Cedeno, Elsayed, Nakov,
  Da~San~Martino, Hasanain, Suwaileh, and Haouari}]{barron2020checkthat}
Alberto Barr{\'o}n-Cedeno, Tamer Elsayed, Preslav Nakov, Giovanni
  Da~San~Martino, Maram Hasanain, Reem Suwaileh, and Fatima Haouari. 2020.
\newblock Checkthat! at clef 2020: Enabling the automatic identification and
  verification of claims in social media.
\newblock In \emph{European Conference on Information Retrieval}, pages
  499--507. Springer.

\bibitem[{Chan et~al.(2021)Chan, Ong, Pung, Zhang, and Fu}]{chan2021cocon}
Alvin Chan, Yew-Soon Ong, Bill Pung, Aston Zhang, and Jie Fu. 2021.
\newblock \href {https://openreview.net/forum?id=VD_ozqvBy4W} {Cocon: A
  self-supervised approach for controlled text generation}.
\newblock In \emph{International Conference on Learning Representations}.

\bibitem[{Devlin et~al.(2019)Devlin, Chang, Lee, and
  Toutanova}]{devlin-etal-2019-bert}
Jacob Devlin, Ming-Wei Chang, Kenton Lee, and Kristina Toutanova. 2019.
\newblock \href {https://doi.org/10.18653/v1/N19-1423} {{BERT}: Pre-training of
  deep bidirectional transformers for language understanding}.
\newblock In \emph{Proceedings of the 2019 Conference of the North {A}merican
  Chapter of the Association for Computational Linguistics: Human Language
  Technologies, Volume 1 (Long and Short Papers)}, pages 4171--4186,
  Minneapolis, Minnesota. Association for Computational Linguistics.

\bibitem[{Fan et~al.(2020)Fan, Piktus, Petroni, Wenzek, Saeidi, Vlachos,
  Bordes, and Riedel}]{fan-etal-2020-generating}
Angela Fan, Aleksandra Piktus, Fabio Petroni, Guillaume Wenzek, Marzieh Saeidi,
  Andreas Vlachos, Antoine Bordes, and Sebastian Riedel. 2020.
\newblock \href {https://doi.org/10.18653/v1/2020.emnlp-main.580} {Generating
  fact checking briefs}.
\newblock In \emph{Proceedings of the 2020 Conference on Empirical Methods in
  Natural Language Processing (EMNLP)}, pages 7147--7161, Online. Association
  for Computational Linguistics.

\bibitem[{Holtzman et~al.(2019)Holtzman, Buys, Du, Forbes, and
  Choi}]{holtzman2019curious}
Ari Holtzman, Jan Buys, Li~Du, Maxwell Forbes, and Yejin Choi. 2019.
\newblock The curious case of neural text degeneration.
\newblock \emph{arXiv preprint arXiv:1904.09751}.

\bibitem[{Keskar et~al.(2019)Keskar, McCann, Varshney, Xiong, and
  Socher}]{keskar2019ctrl}
Nitish~Shirish Keskar, Bryan McCann, Lav~R Varshney, Caiming Xiong, and Richard
  Socher. 2019.
\newblock Ctrl: A conditional transformer language model for controllable
  generation.
\newblock \emph{arXiv preprint arXiv:1909.05858}.

\bibitem[{Lewis et~al.(2019)Lewis, Liu, Goyal, Ghazvininejad, Mohamed, Levy,
  Stoyanov, and Zettlemoyer}]{lewis2019bart}
Mike Lewis, Yinhan Liu, Naman Goyal, Marjan Ghazvininejad, Abdelrahman Mohamed,
  Omer Levy, Ves Stoyanov, and Luke Zettlemoyer. 2019.
\newblock Bart: Denoising sequence-to-sequence pre-training for natural
  language generation, translation, and comprehension.
\newblock \emph{arXiv preprint arXiv:1910.13461}.

\bibitem[{Mansour et~al.(2022)Mansour, Elsayed, and Al-Ali}]{mansour2022did}
Watheq Mansour, Tamer Elsayed, and Abdulaziz Al-Ali. 2022.
\newblock Did i see it before? detecting previously-checked claims over
  twitter.
\newblock In \emph{European Conference on Information Retrieval}, pages
  367--381. Springer.

\bibitem[{Mishra et~al.(2020)Mishra, Gupta, and
  Leippold}]{mishra2020generating}
Rahul Mishra, Dhruv Gupta, and Markus Leippold. 2020.
\newblock Generating fact checking summaries for web claims.
\newblock In \emph{EMNLP W-NUT 2020: Conference on Empirical Methods in Natural
  Language Processing (EMNLP)}.

\bibitem[{Nakov et~al.(2022)Nakov, Barr{\'o}n-Cede{\~n}o, Da~San~Martino, Alam,
  Stru{\ss}, Mandl, M{\'\i}guez, Caselli, Kutlu, Zaghouani
  et~al.}]{nakov2022clef}
Preslav Nakov, Alberto Barr{\'o}n-Cede{\~n}o, Giovanni Da~San~Martino, Firoj
  Alam, Julia~Maria Stru{\ss}, Thomas Mandl, Rub{\'e}n M{\'\i}guez, Tommaso
  Caselli, Mucahid Kutlu, Wajdi Zaghouani, et~al. 2022.
\newblock The clef-2022 checkthat! lab on fighting the covid-19 infodemic and
  fake news detection.
\newblock In \emph{European Conference on Information Retrieval}, pages
  416--428. Springer.

\bibitem[{Nakov et~al.(2021{\natexlab{a}})Nakov, Corney, Hasanain, Alam,
  Elsayed, Barr'on-Cedeno, Papotti, Shaar, and Martino}]{Nakov2021AutomatedFF}
Preslav Nakov, David Corney, Maram Hasanain, Firoj Alam, Tamer Elsayed, Alberto
  Barr'on-Cedeno, Paolo Papotti, Shaden Shaar, and Giovanni Da~San Martino.
  2021{\natexlab{a}}.
\newblock Automated fact-checking for assisting human fact-checkers.
\newblock In \emph{IJCAI}.

\bibitem[{Nakov et~al.(2021{\natexlab{b}})Nakov, Da~San~Martino, Elsayed,
  Barr{\'o}n-Cedeno, M{\'\i}guez, Shaar, Alam, Haouari, Hasanain, Babulkov
  et~al.}]{nakov2021clef}
Preslav Nakov, Giovanni Da~San~Martino, Tamer Elsayed, Alberto
  Barr{\'o}n-Cedeno, Rub{\'e}n M{\'\i}guez, Shaden Shaar, Firoj Alam, Fatima
  Haouari, Maram Hasanain, Nikolay Babulkov, et~al. 2021{\natexlab{b}}.
\newblock The clef-2021 checkthat! lab on detecting check-worthy claims,
  previously fact-checked claims, and fake news.
\newblock In \emph{European Conference on Information Retrieval}, pages
  639--649. Springer.

\bibitem[{Nallapati et~al.(2016)Nallapati, Zhou, dos Santos,
  G$\dot{u}$l{\c{c}}ehre, and Xiang}]{nallapati-etal-2016-abstractive}
Ramesh Nallapati, Bowen Zhou, Cicero dos Santos, {\c{C}}a{\u{g}}lar
  G$\dot{u}$l{\c{c}}ehre, and Bing Xiang. 2016.
\newblock \href {https://doi.org/10.18653/v1/K16-1028} {Abstractive text
  summarization using sequence-to-sequence {RNN}s and beyond}.
\newblock In \emph{Proceedings of The 20th {SIGNLL} Conference on Computational
  Natural Language Learning}, pages 280--290, Berlin, Germany. Association for
  Computational Linguistics.

\bibitem[{Papineni et~al.(2002)Papineni, Roukos, Ward, and
  Zhu}]{papineni-etal-2002-bleu}
Kishore Papineni, Salim Roukos, Todd Ward, and Wei-Jing Zhu. 2002.
\newblock \href {https://doi.org/10.3115/1073083.1073135} {{B}leu: a method for
  automatic evaluation of machine translation}.
\newblock In \emph{Proceedings of the 40th Annual Meeting of the Association
  for Computational Linguistics}, pages 311--318, Philadelphia, Pennsylvania,
  USA. Association for Computational Linguistics.

\bibitem[{Radford and Narasimhan(2018)}]{Radford2018ImprovingLU}
Alec Radford and Karthik Narasimhan. 2018.
\newblock Improving language understanding by generative pre-training.
\newblock In \emph{Published by OpenAI}.

\bibitem[{Radford and Wu(2019)}]{radford2019rewon}
Alec Radford and Jeffrey Wu. 2019.
\newblock Rewon child, david luan, dario amodei, and ilya sutskever. 2019.
\newblock \emph{Language models are unsupervised multitask learners. OpenAI
  Blog}, 1(8):9.

\bibitem[{Raffel et~al.(2020)Raffel, Shazeer, Roberts, Lee, Narang, Matena,
  Zhou, Li, and Liu}]{JMLR:v21:20-074}
Colin Raffel, Noam Shazeer, Adam Roberts, Katherine Lee, Sharan Narang, Michael
  Matena, Yanqi Zhou, Wei Li, and Peter~J. Liu. 2020.
\newblock \href {http://jmlr.org/papers/v21/20-074.html} {Exploring the limits
  of transfer learning with a unified text-to-text transformer}.
\newblock \emph{Journal of Machine Learning Research}, 21(140):1--67.

\bibitem[{Sathe et~al.(2020)Sathe, Ather, Le, Perry, and
  Park}]{sathe-etal-2020-automated}
Aalok Sathe, Salar Ather, Tuan~Manh Le, Nathan Perry, and Joonsuk Park. 2020.
\newblock \href {https://aclanthology.org/2020.lrec-1.849} {Automated
  fact-checking of claims from {W}ikipedia}.
\newblock In \emph{Proceedings of the 12th Language Resources and Evaluation
  Conference}, pages 6874--6882, Marseille, France. European Language Resources
  Association.

\bibitem[{Schuster et~al.(2021)Schuster, Fisch, and
  Barzilay}]{schuster-etal-2021-get}
Tal Schuster, Adam Fisch, and Regina Barzilay. 2021.
\newblock \href {https://doi.org/10.18653/v1/2021.naacl-main.52} {Get your
  vitamin {C}! robust fact verification with contrastive evidence}.
\newblock In \emph{Proceedings of the 2021 Conference of the North American
  Chapter of the Association for Computational Linguistics: Human Language
  Technologies}, pages 624--643, Online. Association for Computational
  Linguistics.

\bibitem[{Shaar et~al.(2021{\natexlab{a}})Shaar, Alam, Martino, and
  Nakov}]{shaar2021assisting}
Shaden Shaar, Firoj Alam, Giovanni Da~San Martino, and Preslav Nakov.
  2021{\natexlab{a}}.
\newblock Assisting the human fact-checkers: Detecting all previously
  fact-checked claims in a document.
\newblock \emph{arXiv preprint arXiv:2109.07410}.

\bibitem[{Shaar et~al.(2021{\natexlab{b}})Shaar, Alam, Martino, and
  Nakov}]{shaar2021role}
Shaden Shaar, Firoj Alam, Giovanni Da~San Martino, and Preslav Nakov.
  2021{\natexlab{b}}.
\newblock The role of context in detecting previously fact-checked claims.
\newblock \emph{arXiv preprint arXiv:2104.07423}.

\bibitem[{Shaar et~al.(2020)Shaar, Babulkov, Da~San~Martino, and
  Nakov}]{shaar-etal-2020-known}
Shaden Shaar, Nikolay Babulkov, Giovanni Da~San~Martino, and Preslav Nakov.
  2020.
\newblock \href {https://doi.org/10.18653/v1/2020.acl-main.332} {That is a
  known lie: Detecting previously fact-checked claims}.
\newblock In \emph{Proceedings of the 58th Annual Meeting of the Association
  for Computational Linguistics}, pages 3607--3618, Online. Association for
  Computational Linguistics.

\bibitem[{Shaar et~al.(2021{\natexlab{c}})Shaar, Haouari, Mansour, Hasanain,
  Babulkov, Alam, Da~San~Martino, Elsayed, and
  Nakov}]{clef-checkthat:2021:task2}
Shaden Shaar, Fatima Haouari, Watheq Mansour, Maram Hasanain, Nikolay Babulkov,
  Firoj Alam, Giovanni Da~San~Martino, Tamer Elsayed, and Preslav Nakov.
  2021{\natexlab{c}}.
\newblock \href {http://ceur-ws.org/Vol-2936/paper-29.pdf} {Overview of the
  {CLEF}-2021 {CheckThat}! lab task 2 on detecting previously fact-checked
  claims in tweets and political debates}.
\newblock In \emph{Working Notes of CLEF 2021---Conference and Labs of the
  Evaluation Forum}, CLEF~'2021, Bucharest, Romania (online).

\bibitem[{Shahi et~al.(2021)Shahi, Dirkson, and Majchrzak}]{SHAHI2021100104}
Gautam~Kishore Shahi, Anne Dirkson, and Tim~A. Majchrzak. 2021.
\newblock \href {https://doi.org/https://doi.org/10.1016/j.osnem.2020.100104}
  {An exploratory study of covid-19 misinformation on twitter}.
\newblock \emph{Online Social Networks and Media}, 22:100104.

\bibitem[{Sharma et~al.(2019)Sharma, Qian, Jiang, Ruchansky, Zhang, and
  Liu}]{sharma2019combating}
Karishma Sharma, Feng Qian, He~Jiang, Natali Ruchansky, Ming Zhang, and Yan
  Liu. 2019.
\newblock Combating fake news: A survey on identification and mitigation
  techniques.
\newblock \emph{ACM Transactions on Intelligent Systems and Technology (TIST)},
  10(3):1--42.

\bibitem[{Shleifer and Rush(2020)}]{shleifer2020pre}
Sam Shleifer and Alexander~M Rush. 2020.
\newblock Pre-trained summarization distillation.
\newblock \emph{arXiv preprint arXiv:2010.13002}.

\bibitem[{Stammbach and Ash(2020)}]{stammbach2020fever}
Dominik Stammbach and Elliott Ash. 2020.
\newblock e-fever: Explanations and summaries for automated fact checking.
\newblock \emph{Proceedings of the 2020 Truth and Trust Online (TTO 2020)},
  pages 32--43.

\bibitem[{Tchechmedjiev et~al.(2019)Tchechmedjiev, Fafalios, Boland, Gasquet,
  Zloch, Zapilko, Dietze, and Todorov}]{tchechmedjiev2019claimskg}
Andon Tchechmedjiev, Pavlos Fafalios, Katarina Boland, Malo Gasquet,
  Matth{\"a}us Zloch, Benjamin Zapilko, Stefan Dietze, and Konstantin Todorov.
  2019.
\newblock Claimskg: a knowledge graph of fact-checked claims.
\newblock In \emph{International Semantic Web Conference}, pages 309--324.
  Springer.

\bibitem[{Thorne et~al.(2018)Thorne, Vlachos, Christodoulopoulos, and
  Mittal}]{thorne-etal-2018-fever}
James Thorne, Andreas Vlachos, Christos Christodoulopoulos, and Arpit Mittal.
  2018.
\newblock \href {https://doi.org/10.18653/v1/N18-1074} {{FEVER}: a large-scale
  dataset for fact extraction and {VER}ification}.
\newblock In \emph{Proceedings of the 2018 Conference of the North {A}merican
  Chapter of the Association for Computational Linguistics: Human Language
  Technologies, Volume 1 (Long Papers)}, pages 809--819, New Orleans,
  Louisiana. Association for Computational Linguistics.

\bibitem[{Vo and Lee(2020)}]{vo-lee-2020-facts}
Nguyen Vo and Kyumin Lee. 2020.
\newblock \href {https://doi.org/10.18653/v1/2020.emnlp-main.621} {Where are
  the facts? searching for fact-checked information to alleviate the spread of
  fake news}.
\newblock In \emph{Proceedings of the 2020 Conference on Empirical Methods in
  Natural Language Processing (EMNLP)}, pages 7717--7731, Online. Association
  for Computational Linguistics.

\bibitem[{Weidinger et~al.(2021)Weidinger, Mellor, Rauh, Griffin, Uesato,
  Huang, Cheng, Glaese, Balle, Kasirzadeh, Kenton, Brown, Hawkins, Stepleton,
  Biles, Birhane, Haas, Rimell, Hendricks, Isaac, Legassick, Irving, and
  Gabriel}]{DBLP:journals/corr/abs-2112-04359}
Laura Weidinger, John Mellor, Maribeth Rauh, Conor Griffin, Jonathan Uesato,
  Po{-}Sen Huang, Myra Cheng, Mia Glaese, Borja Balle, Atoosa Kasirzadeh, Zac
  Kenton, Sasha Brown, Will Hawkins, Tom Stepleton, Courtney Biles, Abeba
  Birhane, Julia Haas, Laura Rimell, Lisa~Anne Hendricks, William~S. Isaac,
  Sean Legassick, Geoffrey Irving, and Iason Gabriel. 2021.
\newblock \href {http://arxiv.org/abs/2112.04359} {Ethical and social risks of
  harm from language models}.
\newblock \emph{CoRR}, abs/2112.04359.

\bibitem[{Xiao et~al.(2021)Xiao, Zhang, Li, Sun, Tian, Wu, and
  Wang}]{10.5555/3491440.3491993}
Dongling Xiao, Han Zhang, Yukun Li, Yu~Sun, Hao Tian, Hua Wu, and Haifeng Wang.
  2021.
\newblock Ernie-gen: An enhanced multi-flow pre-training and fine-tuning
  framework for natural language generation.
\newblock In \emph{Proceedings of the Twenty-Ninth International Joint
  Conference on Artificial Intelligence}, IJCAI'20.

\bibitem[{Zhang et~al.(2020)Zhang, Zhao, Saleh, and Liu}]{zhang2020pegasus}
Jingqing Zhang, Yao Zhao, Mohammad Saleh, and Peter Liu. 2020.
\newblock Pegasus: Pre-training with extracted gap-sentences for abstractive
  summarization.
\newblock In \emph{International Conference on Machine Learning}, pages
  11328--11339. PMLR.

\bibitem[{Zhu et~al.(2019)Zhu, Wang, Wang, Zhou, Zhang, Wang, and
  Zong}]{zhu-etal-2019-ncls}
Junnan Zhu, Qian Wang, Yining Wang, Yu~Zhou, Jiajun Zhang, Shaonan Wang, and
  Chengqing Zong. 2019.
\newblock \href {https://doi.org/10.18653/v1/D19-1302} {{NCLS}: Neural
  cross-lingual summarization}.
\newblock In \emph{Proceedings of the 2019 Conference on Empirical Methods in
  Natural Language Processing and the 9th International Joint Conference on
  Natural Language Processing (EMNLP-IJCNLP)}, pages 3054--3064, Hong Kong,
  China. Association for Computational Linguistics.

\end{thebibliography}
\bibliographystyle{acl_natbib}

\end{document}